\documentclass[runningheads]{llncs}
\usepackage[T1]{fontenc}
\usepackage{graphicx}
\usepackage{amsmath}
\usepackage{booktabs}
\usepackage{multirow}
\usepackage[misc]{ifsym}
\begin{document}
\title{ACA-Net: Future Graph Learning for Logistical Demand-Supply Forecasting}
%
%
\author{
Jiacheng Shi\inst{1} \and
Haibin Wei\inst{1}(\Letter)  \and
Jiang Wang\inst{1} \and
Xiaowei Xu\inst{1} \and
Longzhi Du\inst{1} \and
Taixu Jiang\inst{1}
}
\authorrunning{S. Author et al.}
%
\institute{Alibaba Group, Shanghai, China \\
\email{haibinwei.whb@alibaba-inc.com}
}

\maketitle              
\begin{abstract}
Logistical demand-supply forecasting that evaluates the alignment between projected supply and anticipated demand, is essential for the efficiency and quality of on-demand food delivery platforms and serves as a key indicator for scheduling decisions.  Future order distribution information, which reflects the distribution of orders in on-demand food delivery, is crucial for the performance of logistical demand-supply forecasting. Current studies utilize spatial-temporal analysis methods to model future order distribution information from serious time slices. However, learning future  order distribution in online delivery platform is a time-series-insensitive problem with strong randomness. These approaches often struggle to effectively capture this information while remaining efficient. This paper proposes an innovative spatiotemporal learning model that utilizes only two graphs (ongoing and global) to learn future order distribution information, achieving superior performance compared to traditional spatial-temporal long-series methods. The main contributions are as follows: (1) The introduction of ongoing and global graphs in logistical demand-supply pressure forecasting compared to traditional long time series significantly enhances forecasting performance. (2) An innovative graph learning network framework using adaptive future graph learning and innovative cross attention mechanism (ACA-Net) is proposed to extract future order distribution information, effectively learning a robust future graph that substantially improves logistical demand-supply pressure forecasting outcomes. (3) The effectiveness of the proposed method is validated in real-world production environments. 

\end{abstract}
\section{Introduction}

\begin{figure}
\includegraphics[width=\textwidth]{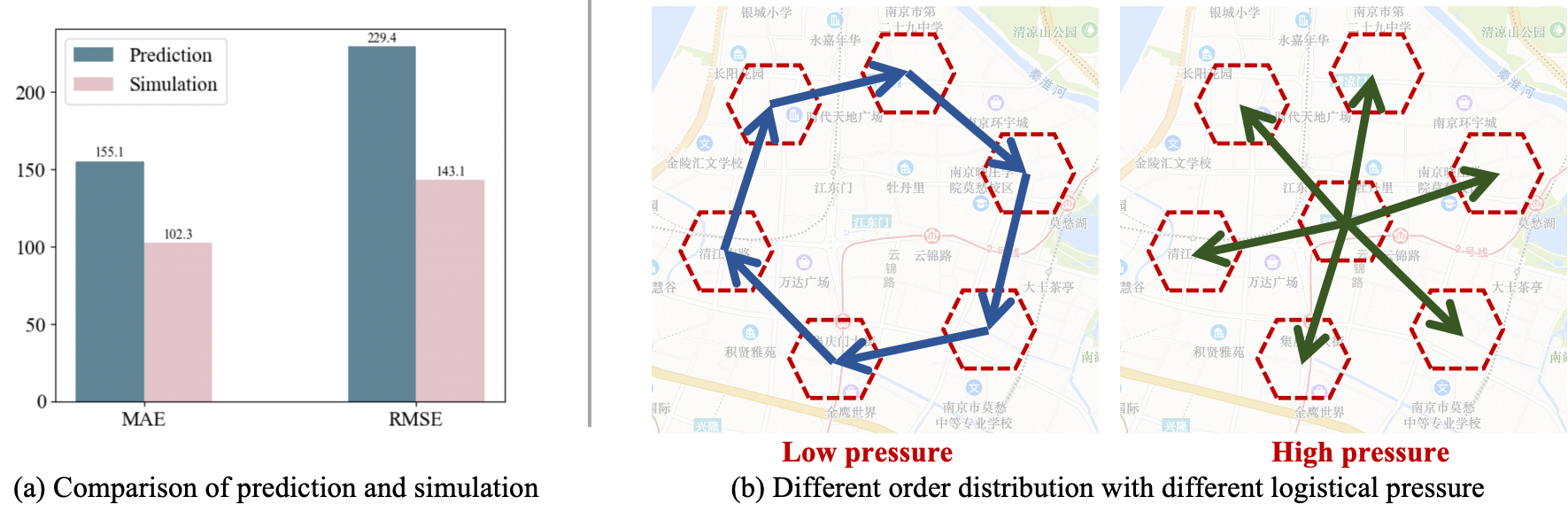}
\caption{(a) It shows the logistical demand-supply forecasting performance of the prediction model and  simulation model. (b) It shows annular and astroid order distribution respectively, which has similar order count but different logistical pressure.} \label{fig1}
\end{figure}

The logistical demand-supply  forecasting\cite{ghalehkhondabi2019review} is essential for enhancing the efficiency and quality of on-demand food delivery (OFD) platforms. It reflects the balance between food delivery riders and online food orders, which in turn affects user experience and delivery costs on OFD platforms. In practical business monitoring, platforms will make different scheduling decisions based on logistical demand-supply forecasting metrics. Hence, the accuracy and compactness of a logistical demand-supply pressure forecasting model are crucial for food delivery logistics dispatchers.

The future order distribution information is essential for improving the accuracy of logistical demand-supply pressure forecasting, which describes the projected distribution of orders during a specified forecast period.  Future order distribution forecasting is anticipated to enhance demand-supply forecasting due to two primary factors: \textbf{a) It  guides the main task.}  Future order distribution forecasting which predicts the future demand distribution is a key part of demand-supply  forecasting. Especially in the context of the OFD, the supply side (represented by riders) is controllable for the platform and allows for access to future rider information, while the demand side (represented by orders)  is more challenging to assess and requires estimation. Furthermore, the order distribution has strong  influence on demand-supply balance as shown in Fig.\ref{fig1} (a) . Therefore, future order distribution forecasting as an auxiliary task steers the demand-supply  forecasting for improved learning efficiency.  \textbf{b) It accelerates the convergence.} Existing researches  \cite{li2022dn,silva2019forecasting}  have proved effective auxiliary tasks will accelerate the training convergence. In this work, we take the advantage of  the predicted future order distribution and our existing simulation model to improve the acceleration.

\begin{figure}
\includegraphics[width=\textwidth]{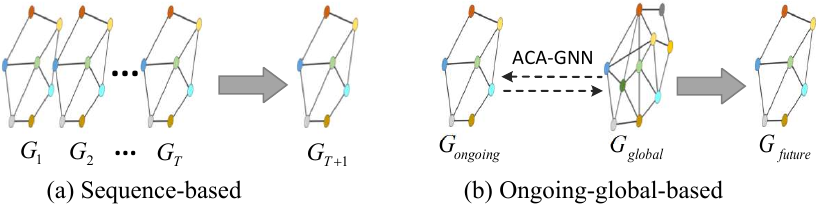}
\caption{Existing sequence-based method and our ongoing-global-based method.} \label{fig2}
\end{figure}

Existing spatiotemporal methods \cite{wang2022hierarchical,wu2019graph}  rely on graphs of past time slices to learn this information.  But these methods fail to meet the accuracy and efficiency requirements of the OFD scenario due to: \textbf{a) Time-series irrelevant and complexity challenge.} Existing spatiotemporal methods\cite{shao2022decoupled,jiang2024sagdfn} learns the predicted future information based on that the future time slice has strong relevance with past time series, as shown in Fig.\ref{fig2} (a). But user ordering has a strong randomness, and its correlation with historical time slices may be at the minute level, or it could be at the daily, monthly, or even yearly level. Therefore, some methods\cite{jin2022deep,DBLP:conf/iclr/LiuYLLLLD22} using long time series of  graphs suggest that by expanding the length of the time series, the issue of randomness in order placement can be mitigated to some extent while it brings complexity challenge.
 \textbf{b) Future graph learning challenge.} Existing methods utilize feature embedding or self-adaptive graph learning to derive the future order distribution information. Feature embedding generates feature representations without robust future graph learning. In comparison, self-adaptive graph learning automatically produces a task-oriented graph\cite{jin2022adaptive}, but this graph lacks reliability. These methods are limited and cannot guarantee the effectiveness and reliability of the learned future graph.

To address the above challenges, we propose a novel future graph learning method called ACA-Net. \textbf{For the time-series irrelevant challenge and complexity challenge:}  As shown in Fig.\ref{fig2} (b), we introduce two types of graph information: the ongoing graph and the global graph. The ongoing graph represents the current order conditions, while the global graph captures historical order statistics. These graphs replace traditional long time series graphs and demonstrate improved performance. \textbf{For the future graph learning challenge:} We propose a novel network ACA-Net to learn the future graph from the ongoing graph and global graph using adaptive graph learning (A) and cross-attention mechanism (CA) : (1) A cross-attention-based feature encoder that facilitates effective relationship learning. We employ inter-graph cross-attention to derive the future order information from ongoing and global graph, and propose rider-environment cross-attention to learn the influence of rider and environment to the future order distribution. (2) A supervised adaptive graph learning component that ensures the reliability of the future graph. We generate the future graph from the features extracted through cross-attention using a self-adaptive graph learning method. Additionally, an auxiliary supervised loss is incorporated to guarantee the reliability of the learned future graph. \textbf{For better utilization of the learned graph:}  A pre-trained simulation model that models the relationship between pressure and the distribution of supply and demand and the environmental factors is applied in ACA-Net to reduce the difficulty of learning.

After learning the future graph that represents the effective future order distribution information, the network forecasts the final logistical demand-supply pressure based on this future graph. It is evaluated on a real-world dataset and has been implemented in actual production, successfully validating its effectiveness. The ACA-Net is currently deployed on one of the world's largest OFD platforms, which serves millions of meals daily across over 300 cities in China.

\section{Problem Definition}
\textbf{Definition 1: Logistical demand-supply forecasting.}  In the context of the food delivery industry, logistical demand-supply balance refers to the relationship between the delivery capacity of riders and the demand for food orders.  In our industrial practice, we choose \textit{the average delivery time of orders created within the next five minutes} as a specific measure of logistical demand-supply balance for two key advantages: first, delivery time is the most tangible factor experienced by customers, merchants, and riders, significantly impacting user experience; thus, using it to represent logistical demand-supply pressure enhances business interpretation. Second, as the average delivery time serves as a prediction target, labels can be easily obtained during dataset construction. Therefore, the logistical demand-supply forecasting of a given district (our management unit) can be expressed by the following formula:
\\
\begin{equation}
p_{a}=\frac{1}{n} \sum^{n}_{i=1} t_{i}\label{eq1}
\end{equation}
Where $p_{a}$ represents the demand-supply balance value of business district $a$. Furthermore, $n$ the number of orders created in business district $a$ within the next five minutes, and $t_{i}$ is the separate delivery time of each order.
\\
\textbf{Definition 2: Global Graph.}  The Global Graph is a graph-structured dataset organized according to historical order information. We extend the statistical period to its maximum extent. This enables the Global Graph to include all potential order distribution scenarios and grants it a global attribute. The Global Graph for a specific business district $a$ is denoted as $G^{a}_{global}=\left( V_{a},E_{a}\right)$. The nodes $V_{a}=\left\{ V_{a_{1}},V_{a_{2}},...,V_{a_{N}}\right\}$ (where $N$ is the number of the nodes) collectively represent the total collection of origin and destination Areas of Interest (AOIs) for all historical orders. An AOI designates a geographical area that includes a conglomerate of merchants or businesses. The feature matrix of the nodes has dimensions $N\times F_{AOI}$. It represents the $F_{AOI}$ features of each node, such as the merchant's meal preparation speed, consumer delivery difficulty, and order volumes. The edges $E_{a}=\left\{ E_{a_{1}},E_{a_{2}},...,E_{a_{N^{'}}}\right\}$ (where $N^{'}$ is the number of the edges) in the global graph describe the flow of orders and are directed. A directed edge from AOI $x$ to AOI $y$ indicates the flow of orders between them. This edge has attributes that include the average daily order volume and the average delivery time per order.
\\
\textbf{Definition 3: Ongoing Graph.} The ongoing graph describes real-time order distribution, closely aligning with future order conditions. Its structure is similar to global graph. The ongoing graph for a business district $a$ is denoted as $G^{a}_{ongoing}=\left( V^{\prime }_{a},E^{\prime }_{a}\right)  $. The Nodes $V^{\prime }_{a}$ in the ongoing graph share the same feature dimension $F_{AOI}$ as those in the global graph, and it holds that $V^{\prime }_{a} \subseteq  V_{a}$. The edges in the ongoing graph represent real-time order distribution are directed.
\\
\textbf{Definition 4: Supply and Environment Features.} In addition to order distribution, the key factors affecting the demand-supply forecasting of a specific business district include supply distribution and real-time environmental elements such as meteorological conditions, courier availability, and traffic congestion status. Supply and environment features are represented as $F\in R^{1\times N_{f}}$, where $N_{f}$ indicates the number of features.

\begin{figure}[!t]
\includegraphics[width=\textwidth]{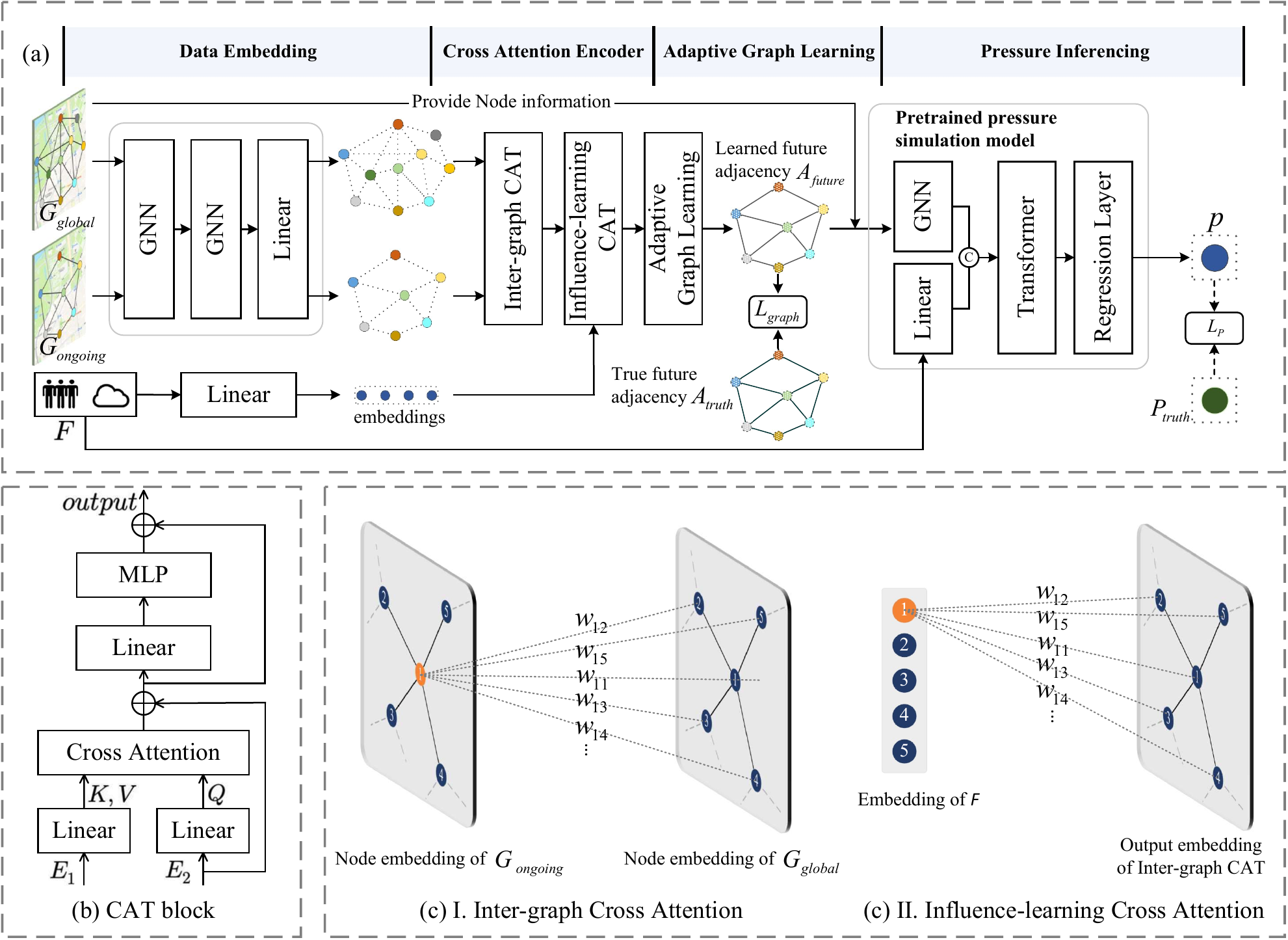}
\caption{The overview of the proposed ACA-Net.} \label{fig4}
\end{figure}

\section{Methodology}
As shown in Fig.\ref{fig4} (a), Our ACA-Net generate the future graph $G_{future}$ from the introduced ongoing and global graph through our proposed ACA-Net networks. ACA-Net is composed of four key components: (a) Data Embedding: This component is responsible for learning the node feature embeddings of both ongoing and global graphs, in addition to the embeddings of rider and environmental features. (b) Cross Attention Encoder: ACA-Net employs inter-graph cross attention to capture the relationships between ongoing and global graphs and utilizes influence-learning cross attention to discern the impact of environmental and supply factors on future order distribution. (c) Adaptive Graph Learning: This component utilizes an auxiliary supervised learning method to generate the future graph from the extracted feature representations. (d) Pressure Inferencing: Leveraging a pre-trained pressure simulation model, this component infers the target pressure ratio based on the learned future graph.

\subsection{Data Embedding}
\textbf{Graph Embedding}
To reduce the dimensions of nodes and edges of the ongoing graph and global graph inputs, we utilize GNNs for graph embedding, thereby mapping the high-dimensional graph structure to a low-dimensional vector space.  As shown in Fig.\ref{fig4} (a), Graph Embedding modules utilize two GNN layers to learn the graph feature and one linear layer to map the node embedding to a low-dimensional vector. The input node features of $G_{ongoing}$ and $G_{global}$ have the size of $M_{on} \times F_{aoi}$ and $M_{glo} \times F_{aoi}$ ,and after the Graph Embedding the output sizes are $m \times C$ and $M \times C$ respectively, where $m$ and $M$ are the node count of the ongoing graph embedding and global graph embedding and $C$ is the feature channel size of two graph embeddings.
\\
\textbf{Supply and environment feature Embedding}
Supply and environment features ($F$) provide the information of rider distribution and environment factor  of the supply and demand pressure. ACA-Net takes basic linear layer as embedding layer to project $F$ to an vector with size of $C_{f}$.

\subsection{Cross Attention Encoder}
\textbf{Cross Attention Transformer (CAT) block.}
The overall structure of Cross Attention Transformer(CAT) block is shown in the Fig.\ref{fig4} (b). It calculates the cross attention between the node embeddings $E_{1}$ and $E_{2}$ , and utilizes a residual connection after the output of cross attention to avoid the common problem of gradient vanishing in deep networks and to enhance the model's learning capability.   $MLP$ donates the Linear layer and Multi-Layer Perceptron which is used to enhance the expressiveness of the model. We proposed two kinds of cross attention named inter-graph cross attention and influence-learning cross attention to learn the relationships between different features.
\\
\textbf{Inter-graph CAT for graph relationship learning.}
The Inter-graph Cross Attention Transformer (CAT) takes advantage of  inter-graph cross attention mechanism to learn the relationship between ongoing graph and global graph.  The calculation process of Inter-graph CAT can be donated as Eq.\ref{eq2} and Eq.\ref{eq3} where $E_{ongoing}$ and $E_{global}$ are the node embedding of $G_{ongoing}$ and $G_{global}$ respectively. $E_{half}$ is the middle output of the inter-graph cross attention.  
\begin{gather}
    E_{half}=E_{ongoing}+Attention(LN(E_{ongoing}),LN(E_{global}))\label{eq2}\\
    E_{on-glo}=E_{half}+MLP(LN(E_{half}))\label{eq3}
\end{gather}

As shown in Fig.\ref{fig4} (c), inter-graph cross attention computes the corresponding weights from the ongoing graph to the global graph. It shows an example node embedding in ongoing graph and it calculates the weights $w_{11}, w_{12}, w_{13}, w_{14}, w_{15}...$ with all the node embeddings in global feature. The weights is calculated as Eq.\ref{eq4} where $ Q, K$ are the $ query, key$ matrix projected from the node embeddings of ongoing graph and global graph and $d$ is the dimension of $query/key$. The output of the cross attention mechanism is the  weighted sum of $weight$ and $value$, where the $ value $ is projected from the global graph. Therefore, the attention is calculated as Eq.\ref{eq5}. 
\begin{equation}
Weight\left( Q,K\right)  =softmax\left( \frac{QK^{T}}{\sqrt{d} } \right) \label{eq4}
\end{equation}  
\begin{equation}
Attention\left( Q,K,V\right)  =softmax\left( \frac{QK^{T}}{\sqrt{d} } \right)  V \label{eq5}
\end{equation}      
The process of generating  $ query, key, value$ from a pair of node embeddings of ongoing graph and global graph  $E_{ongoing}$ and $E_{global}$ can be donated as Eq.\ref{eq6} where $\chi \left( .\right)  $ denotes a function reshaping a feature maps into the desired form and $\left\{ \omega_{q} ,\omega_{k} ,\omega_{v} \right\}  $ represent  linear  projections.
\begin{equation}
\label{eq6}
\begin{split}
&q_{i}=[\chi_{q} \left( E_{ongoing}\right)  ]^{T}\omega_{q} \\ 
&k_{i}=[\chi_{k} \left( E_{global}\right)  ]^{T}\omega_{k} \  \  i\in \left\{ 1,2,...,n\right\}  \\ 
&v_{i}=[\chi_{v} \left( E_{global}\right)  ]^{T}\omega_{v} 
\end{split} 
\end{equation}
\\
\textbf{Influence-learning CAT for supply and environment features.} Supply and environment features, donated as $F$, are crucial influencing factors for future graph learning and demand-supply forecasting. We proposed a cross-attention-based method to learn how the supply and environment features influence the future graph learning. As shown in Fig.\ref{fig4} (b), influence-learning CAT takes the similar structure as Inter-graph CAT and only the cross attention mechanism is different. 

Influence-learning cross attention is illustrated in the Fig.\ref{fig4} (c).  It takes the embedding of $F$ as the $query$ and take the output embedding of inter-graph CAT as the $key$ and $value$. It will learn how each rider-environment feature influence the order distribution information in the graph embedding. 
\subsection{Adaptive Graph Learning (AGL)}
Adaptive graph learning is used to learn the adjacency matrix $A_{future}$ of future graph automatically. $A_{future}$ is generated from the extracted feature embedding $E_{out}$ from influence-learning CAT. The calculation formula is as Eq.\ref{eq7}. We derive the future spatial dependency weights by multiplying the  $E_{out}$ and itself. The LN is a linear layer to project the feature to the target shape. The ReLU activation function is used to eliminate weak connections. The SoftMax function is applied to normalize the self-adaptive adjacency matrix. The normalized self-adaptive adjacency matrix, therefore, can be considered as the transition matrix of a hidden diffusion process. 
\begin{equation}
A_{future} =SoftMax\left( ReLU\left( LN\left(E_{out}E^{T}_{out}\right) \right)  \right) \label{eq7} 
\end{equation}

In order to ensure that the learned future graph can fit the distribution of future real orders, we have adopted a supervised approach to constrain the learning of the future graph. As shown in Fig.\ref{fig4} (a), A loss function $L_{graph}$ is utilized to constrain the difference between  the learned  adjacency matrix $A_{future}$ of future graph and the ground truth adjacency matrix $A_{truth}$. In our graph structure, the edge has two different attributes including delivery time and order count, and the elements $A_{truth}$ are obtained by multiplying the two aforementioned attributes. The $L_{graph}$ is calculated as:
\begin{equation}
L_{graph} = MSE(A_{future},A_{truth})\label{eq8}
\end{equation}
\subsection{Pressure Inferencing}
\textbf{Pre-trained simulation model.} The logistical supply and demand pressure is accurately obtained by a simulation model if the all the supply, demand and environment information are achievable. Therefore, we takes the advantage of  a pre-trained pressure simulation model to reduce the difficulty of the forecasting model. The pre-trained dataset is composed of the demand distribution, the supply distribution, the environment factors and the corresponding target index in 5 minutes.  The demand distribution is represented by graph data for detailed and effective description, and other features are represented by a one-dimensional vector because they are predictable.
\\
\textbf{Inferencing layers based on future graph.}We suppose that the global graph can provide information on all nodes that may appear in the future. Therefore, after obtaining the edge information $A_{future}$ of the future graph and combining it with the node information provided by the global graph, we can obtain complete information on the future graph. Therefore, the learned $G_{future}$ and the input $F$ provides all the supply, demand and environment information. The output pressure $p$ through the  simulation model  $f_{2}\left( \cdot \right)  $  is donated as:
\begin{equation}
p=f_{2}\left( G_{future},F\right)  \label{eq9}
\end{equation}
\\
\textbf{Loss function.} In this paper, we use Mean Absolute Error (MAE) of real value and predicted value as the loss function, which can be expressed as Eq.\ref{eq10} and the overall loss function of our ACA-Net is calculated as Eq.\ref{eq11} where the $\lambda$ is the weight of  $L_{graph}$. By adjusting the value of $\lambda$, the importance of the future graph learning loss throughout the task can be controlled, thereby ensuring that the ACA-Net can not only learn an effective future graph but also achieve excellent pressure forecasting performance.
\begin{equation}
L_{P}=MAE\left( Output,\  Truth\right)  =\frac{1}{Q} \sum^{t=P+Q}_{t=P+1} \left| Y_{t}-\hat{Y}_{t} \right|  \label{eq10}
\end{equation}
\begin{equation}
L=L_{P}+\lambda L_{graph} \label{eq11}
\end{equation}

\begin{table}
\caption{It shows the SOTA performance of our proposed ACA-Net compared with other methods from the point of accuracy and lightweightness.}\label{tab2}
\begin{center}
\setlength{\tabcolsep}{8pt}
\begin{tabular}{ccccccc}
\toprule
\multirow{2.5}{*}{\textbf{Methods}}&\multicolumn{3}{c}{\textbf{Accuracy}}&\multicolumn{2}{c}{\textbf{Lightweightness}}\\
\cmidrule(r){2-4} \cmidrule(r){5-6} 
  &  MAE & RMSE & MAPE & run time & input bytes \\
\midrule
XGB\cite{chen2016xgboost} & $163.6$ & $241.3$ & $0.134$ & $0.021$ &$3.0 \times 10^3$  \\

DeepFM\cite{guo2017deepfm} & $155.1$ & $229.4$ & $0.123$ & $0.992$ &$3.0 \times 10^3$ \\

LSTM\cite{abbasimehr2020optimized} & $149.6$ & $211.3$ & $0.105$ & $1.582$ &$3.0 \times 10^4$ \\

Informer\cite{zhou2021informer} & $146.3$ & $206.4$ & $0.101$ & $2.599$ &$3.0 \times 10^4$ \\

STGNN\cite{wang2020traffic} & $138.1$ & $186.3$ & $0.089$ & $4.134$ &$1.5 \times 10^7$ \\

ASTGNN\cite{guo2021learning} & $136.1$ & $183.3$ & $0.085$ & $6.283$ &$1.5 \times 10^7$ \\

D2STGNN\cite{shao2022decoupled} & $136.9$ & $182.3$ & $0.081$ & $5.217$ &$1.5 \times 10^7$ \\

SAGDFN\cite{jiang2024sagdfn} & $135.1$ & $181.7$ & $0.079$ & $3.950$ &$1.5 \times 10^7$ \\

\textbf{(our) ACA-Net}  & $\textbf{126.3}$ & $\textbf{169.7}$ & $\textbf{0.071}$ & $3.206$ &$1.5 \times 10^6$ \\
\bottomrule
\end{tabular}
\end{center}
\end{table}

\section{Experiment}
\subsection{Experimental Settings}

\textbf{Datasets description.}  Our data is collected and organized from one month of real delivery data from our food delivery platform. We leverage data from two typical cities, Shanghai and Nanjing, to assess the performance of our proposed method. Each data point reflects the food delivery supply and demand situation in a specific business district at a certain minute. The dataset consists of a total of 565,130 data records, including 496,917 training records, 12,408 validation records, and 52,805 test records. All datasets are preprocessed according to business requirements to ensure that the pressure ratio signal can effectively guide online control measures.
\\
\textbf{Evaluation metrics.} In this paper, we focus on both the accuracy and lightweight nature of the model because these metrics will influence online production.  In the experiments, MAE, RMSE, and MAPE are used to assess the model's accuracy, while runtime and input bytes are utilized to evaluate its lightweight nature. Runtime refers to the time consumed by the model to infer a batch, where the batch size remains consistent across all experiments. Input bytes represent the size of the input data for the model which is related to the time required for collecting data online. 
\\
\textbf{Baselines models.} We compare our proposed approach with the following baseline methods: \textit{DeepFM}\cite{guo2017deepfm} is a benchmark factorization model for deep learning, proven effective in many regression tasks. The input features of DeepFM are the statistical features at time slice $t$.  \textit{XGB}\cite{chen2016xgboost}  is a basic tree model that represents the traditional machine learning methods without deep learning. The input features of XGB are the same as DeepFM. \textit{LSTM}\cite{abbasimehr2020optimized} and \textit{Informer}\cite{zhou2021informer} are two classic sequence models for time-series features. Their input features consist of the statistical features from 10 time slices between $t-\Delta t$ and $t$. \textit{STGNN}\cite{wang2020traffic}, \textit{ASTGNN}\cite{guo2021learning}, \textit{D2STGNN}\cite{shao2022decoupled}, \textit{SAGDFN}\cite{jiang2024sagdfn}  are spatio-temporal models that leverage both spatial and temporal features. They utilize statistical features and the continuous order distribution graphs of past $\Delta t$ time slices to predict the logistical demand-supply pressure at $t+1$. 
\\
\textbf{parameter settings.} The learning rate is set as $0.001$. All numeric features undergo z-score normalization, except for the edge features of the graphs, which utilize Min-Max normalization to ensure that their values are greater than zero. In ACA-Net, we implement multi-head cross attention with the number of heads set to $H=8$. The output node embedding sizes are $m=100$ and $M=500$. The weight in the loss function is $\lambda=0.1$.

\begin{figure}[!t]
\includegraphics[width=\textwidth]{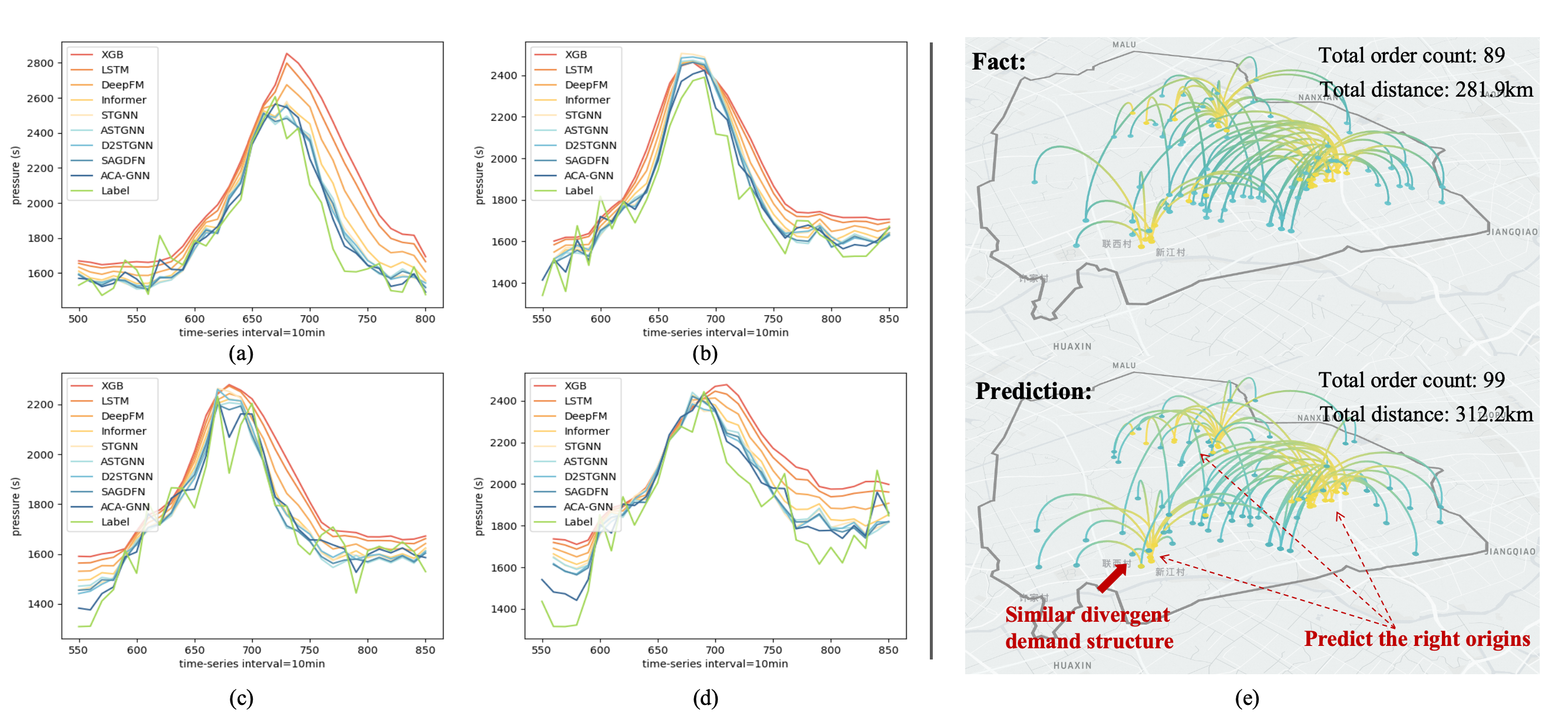}
\caption{(a)(b)(c)(d) Result curves of ACA-Net and contrast experiments. (e) Visualization of the predicted future graph and the real graph.} \label{fig7}
\end{figure}

\subsection{Performance comparison}
Tab.\ref{tab2} and Fig.\ref{fig7}(a)(b)(c)(d) show the performance of our model alongside contrast experiments. The performance of XGB is poor when compared with other deep learning methods illustrating deep learning approaches are more suitable for addressing complex prediction problems. Both DeepFM and XGB utilize only numeric features without graphs, which perform worse than methods incorporating spatial or temporal features. Therefore, spatial and temporal features are crucial for our logistical demand-supply pressure forecasting task. LSTM and Informer primarily leverage time-series features. However, they show only a minor improvement in performance compared with methods that do not include temporal features. Our proposed ACA-Net outperforms both LSTM and Informer. Although time-series features are not explicitly input into the model, ACA-Net effectively learns useful temporal features through ongoing and global graphs. STGNN and its variants, including ASTGNN, D2STGNN, and SAGDFN, achieve exceptional performance on logistical demand-supply pressure forecasting tasks. However, these models exhibit higher resource consumption (5 times the input bytes) and slower inference speeds compared to our ACA-Net. Remarkably, our ACA-Net achieves the best performance among all implemented methods while using only one graph with two different adjacency matrices (ongoing and global).

\subsection{Ablation study}
Ablation experiments are conducted by masking components of ACA-Net to demonstrate the importance of each innovation. As shown in Tab.\ref{tab3}, MAE serves as the benchmark for improvement. Adding the ongoing and global graphs results in the performance enhancement which illustrates the effectiveness of the introducing graphs. The cross attention mechanism promotes the feature learning ability of the proposed ongoing and global graphs and brings obvious improvement. As shown in Fig.\ref{fig7}(e), the learned future graph through ACA-Net is similar to the real graph whose total order count and distance are close, illustrating the effective graph learning of  proposed network. The incorporation of adaptive graph learning to predict future graph structures, alongside the utilization of a simulation model, significantly enhances the training efficacy of our model. These advancements are pivotal in explaining why our proposed ACA-Net outperforms traditional spatiotemporal methods.

\begin{table}
\caption{Ablation of our innovations.}\label{tab3}
\begin{center}
\begin{tabular}{ccccc}
\hline
Ongoing graph &  Global graph & Cross Attention & Adaptive learning & MAE\\
\hline
$ $ &  &  & &$155.1$\\
$\surd$ &  & & &$152.2$\\
 & $\surd$ &  & &$148.1$\\
  $\surd$ & $\surd$ &  & &$145.8$\\
$\surd$ & $\surd$ & $\surd$ & &$136.4$\\
$\surd$ & $\surd$ & $\surd$ &$\surd$ &$\textbf{126.3}$\\
\hline
\end{tabular}
\end{center}
\end{table}

\section{Conclusion}
This paper proposes a model named ACA-Net that effectively learns future information while utilizing fewer input time slices for the online production industry on the OFD platform. Furthermore, ACA-Net generates an explainable and robust future graph and utilizes a simulation model for better performance.  This method has been implemented online and achieved significant business gains.

\subsubsection{Acknowledgements.} This work is supported by Ele.me, which is one of the largest online food delivery platforms in China. We sincerely appreciate the collaborative efforts of the Ele.me Logistics Technology Department and the cloud computing resources provided by Alibaba.

%
%
%
\bibliographystyle{splncs04}
\bibliography{references}
\end{document}